\documentclass[11pt]{article}

\usepackage[final]{acl}

\usepackage{times}
\usepackage{latexsym}

\usepackage[T1]{fontenc}
\usepackage[utf8]{inputenc}

\usepackage{microtype}

\usepackage{inconsolata}

\usepackage{graphicx}
\usepackage[T1]{fontenc}
\usepackage[utf8]{inputenc}
\usepackage{microtype}
\usepackage{booktabs}
\usepackage{tabularx}
\usepackage{array}
\usepackage{amsmath}
\newcolumntype{Y}{>{\raggedright\arraybackslash}X}
\newcolumntype{P}[1]{>{\raggedright\arraybackslash}p{#1}}

\title{AI-Enabled Quality Assurance for Multiple-Choice Assessment Items}

\author{Steven Moore \\
  George Mason University \\
  \texttt{stevenjamesmoore@gmail.com} \\\And
  Nicholas Diana \\
  Colgate University \\
  \texttt{ndiana@colgate.edu} \\}

\begin{document}
\maketitle

\begin{abstract}
Generating multiple-choice questions is increasingly scalable, but establishing their assessment quality remains difficult. We present a focused narrative review of automated item-writing flaw detection, revision, psychometric screening, and NLP benchmark auditing. Database searches, citation retrieval, and nominated sources yield fourteen research reports reviewed in full text. We distinguish surface checks from content-sensitive judgments and map a 19-criterion rubric to detection methods and reported evidence. High label-level accuracy often coexists with weak positive-case detection, while rubric definitions and reference standards vary. Revision evidence is mixed, and the associations reported in prior work do not establish the effects of repair. We propose evaluating quality assurance as a sequence of independently validated decisions, with criterion-specific reporting, calibrated human review, and outcome-based assessment of revisions.
\end{abstract}

\section{Introduction}

Multiple-choice questions (MCQs) support efficient assessment but require careful design \citep{gottlieb2023educator}. A distractor may be grammatically inconsistent, implausible, or defensible as a second answer. These problems require different forms of review. Item-writing guidelines provide a vocabulary for identifying them \citep{haladyna2002,tarrant2006}, yet applying it across large item banks remains demanding.

Large language models (LLMs) change the scale of authoring without eliminating the need for validation \citep{balepur2025these}. Results in programming education demonstrate that GPT-4 can generate promising questions aligned with learning objectives \citep{doughty2024}. The relevant next question is how to determine which generated or existing items are suitable for a particular assessment. Textual fluency, agreement with an answer key, and appropriate measurement of knowledge are distinct properties. Reference-overlap metrics such as BLEU and ROUGE measure similarity to reference text, not whether distractors are defensible or a question measures its intended construct~\citep{moore2024}. Human judgments also depend on the evaluation protocol and raters' expertise \citep{vanderlee2021}.

The Item-Writing Flaws (IWF) approach makes quality review more interpretable by assigning named flags to specific problems \citep{tarrant2006}. The Scalable Automatic Question Usability Evaluation Toolkit (SAQUET) operationalizes a 19-criterion version of the IWF rubric using rules, NLP, and selective LLM verification \citep{moore2024}. Related work extends automated review toward revision, psychometric and Item Response Theory (IRT) analysis, and auditing the MCQs used to evaluate language models. These applications share a rubric vocabulary, but make different evidentiary claims.

We contribute a criterion-level map of the automated application of IWFs, a comparison of detection and downstream evidence, and a research agenda connecting flags to assessment decisions. We distinguish \emph{detecting a rubric violation}, \emph{improving an item}, and \emph{improving measurement}, noting that success at one stage does not establish success at the next. In measurement terms, a removed flaw is a change to the item, not yet evidence about the scores it produces.

\section{Review Scope and Method}
\label{sec:method}
We conducted a focused narrative evidence map around five seed reports covering detection, remediation, psychometric and cognitive outcomes, and benchmark auditing. Searches of PubMed and Crossref on September 1, 2026 supplemented this set with records published from January 1, 2020 through that date. PubMed provided access to biomedical and medical education literature on MCQ design and quality evaluation. Crossref provided complementary discovery through multidisciplinary bibliographic metadata, helping identify work beyond biomedicine~\citep{gusenbauer2024beyond}. Targeted searches of education and computing venues and citation-based discovery completed the search. This bounded strategy does not establish exhaustive coverage of the relevant literature.

\begin{samepage}
\noindent The title and abstract query was:
\begin{quote}\small
("multiple choice" OR "multiple-choice") AND ("item writing" OR "item-writing" OR "quality") AND ("large language model" OR "ChatGPT" OR "automated" OR "natural language processing" OR "NLP").
\end{quote}
\end{samepage}
Crossref's \texttt{query.bibliographic} searches used \emph{item-writing flaws} and \emph{multiple choice question quality large language model}, retaining the first 25 relevance-ranked records per query. Exploratory searches targeting ACL Anthology, ACM, Springer, and ERIC, followed by title and DOI lookups, helped locate sources. References in the retrieved papers identified the earlier comparison by \citet{moore2023}.

Eligibility required an empirical evaluation of automated MCQ flaw detection or correction, or an analysis connecting automated flaw annotations to authoring, psychometric, or benchmark outcomes. Generation-only studies, model exam-taking studies without an evaluated flaw-screening component, general reviews, and unrelated assessment formats were excluded from the core synthesis. We included related technical-flaw rubrics without assuming equivalence to all 19 IWF criteria.

The databases returned 212 records, 162 from PubMed and 50 from Crossref. One duplicate DOI and one preprint/published-version pair left 210 reports for screening. Seven research papers met the criteria and 203 were excluded. Adding five seed reports, one citation-linked predecessor~\citep{moore2023}, and one subsequently nominated report~\citep{arif2024} yielded fourteen included reports, all reviewed in full text (Table~\ref{tab:studies}). Extraction distinguished the item sample, reference standard, rubric version, detector, metric denominator, and outcome. We prioritized criterion-level performance, positive class F1, and chance corrected agreement. Missing metrics remained unreported and aggregate accuracy was not interpreted as recall. Dataset reuse and incompatible units precluded summing sample sizes or pooling performance.

\begin{table*}[t]
\centering
\small
\begin{tabularx}{\textwidth}{@{}P{0.225\textwidth}P{0.29\textwidth}Y@{}}
\toprule
\textbf{Report / evidence} & \textbf{Sample and reference standard} & \textbf{What the evaluation establishes} \\
\midrule
\citet{moore2023} & 200 student-authored MCQs; two expert raters & Rules vs. GPT-4; domain micro-F1 .30--.70 vs. .25--.36. \\
\addlinespace
\citet{moore2024} & 271 MCQs, five domains; existing expert labels & SAQUET: label accuracy 94.13\%, exact match 38\%; criterion evidence in Table~\ref{tab:criteria}. \\
\addlinespace
\citet{arif2024} & 1,709 GPT-3.5 and 1,850 GPT-4 MCQs; 100 human-rated items & Rule/LLM rubric audits of video-based questions; missing context and distractor flaws; pass rates are not detector accuracy. \\
\addlinespace
\citet{sabqat2025} & 13 NBME examples, one per flaw category & GPT-3.5 detection/correction before and after guideline prompting; no representative negative-item sample. \\
\addlinespace
\citet{aydin2026} & 85 internal-medicine MCQs; three-expert reference & Three LLMs; negative-stem $\kappa=.897$ for Claude and Gemini; learning-outcome alignment ICC .121--.380. \\
\addlinespace
\citet{gerard2026} & Clinical pharmacology; 87 entries in the diagnostic 2x2 table & Shared model/key disagreement: sensitivity 35.3\%, specificity 92.9\%, precision 54.6\%, NPV 85.5\%. \\
\addlinespace
\citet{iembo2026} & 36 original/revised pairs; 38 criteria; external and internal panels & External ratings found no significant aggregate change; calibrated author ratings favored revised items. \\
\addlinespace
\citet{menze2026} & 152 items, 19 teachers; two LLMs; 30 human-rated items & AI/human-consensus Hamming loss .16; AI-supported items had more rubric flags. \\
\addlinespace
\citet{schmucker2025} & 7,126 STEM MCQs; estimated IRT parameters & Automated flags predict some parameter variation; difficulty $R^2$ reaches .216 overall but .030 in mathematics. \\
\addlinespace
\citet{moore2026cognitive} & 1,230 MCQs, three datasets; item-level response time and difficulty & Confusion-category flags associate with 14.0--29.4\% longer response time; observational, category-level evidence. \\
\addlinespace
\citet{balepur2026} & 12 NLP MCQ benchmarks; human annotation decisions & Rubric judges and benchmark filtering; NLP writing-flaw F1 .66, $\kappa=.53$ for Gemini 2.5 Pro. \\
\addlinespace
\citet{firoozi2025} & 20 macroeconomics MCQs; two item-writing experts and one subject expert & Six-rule/GPT-4o revision: 17/20 flawless, 18/20 linguistically acceptable, 14/20 retain intended content. \\
\addlinespace
\citet{duesterbeck2026} & 30 German biochemistry MCQs; six experts compare two refinements & Optimized agent sequences preferred for stems over a fixed pipeline; no option advantage. \\
\addlinespace
\citet{wu2026} & 43 defective dental MCQs; 44 flaws identified by two experts & GPT-4 detects 12/44 flaws (27.3\%); 7/7 duplicates but 1/23 off-topic flaws; no defect-free controls. \\
\bottomrule
\end{tabularx}
\caption{Evidence map, not a system ranking, with all rows drawing on full text reports. Samples, rubric definitions, model versions, and outcomes differ. NPV is negative predictive value, ICC is intraclass correlation, and reported performance applies to the stated task and reference standard.}
\label{tab:studies}
\end{table*}

\section{What the Rubric Operationalizes}
\label{sec:rubric}

IWF criteria describe avoidable sources of cueing, ambiguity, or irrelevant task demands \citep{tarrant2006}. They do not constitute a complete theory of assessment quality. Content coverage, curricular alignment, fairness, and intended cognitive demand require additional evidence. An easy item may be appropriate for checking prerequisite knowledge, and a difficult item may be valuable for distinguishing advanced performance. Table~\ref{tab:criteria} uses SAQUET's definitions as the reference vocabulary \citep{moore2024}. Its routes illustrate three levels of automation. Text checks address visible patterns or format and NLP adds relationships among words and options. Selective LLM verification attempts judgments for which a pattern alone is insufficient. These routes describe implementation choices, not a fixed ordering of pedagogical importance. For example, absolute terms are easy to locate but may be entirely appropriate in a mathematical statement.

The rubric's lineage should not be mistaken for identical operationalization. \citet{costello2018} describe 15 manually assessed flaws alongside programmatic item characteristics. SAQUET reports different human thresholds for the longest-correct-option criterion across its source datasets \citep{moore2024}. BenchMarker separately represents a focused question and a problem fully stated in the stem, lacks SAQUET's explicit true/false criterion, and broadens the length rule to balanced options \citep{balepur2026}. Several medical evaluations also adapt National Board of Medical Examiners (NBME) guidance \citep{sabqat2025}. Accordingly, we preserve source definitions rather than treating equally named scores as interchangeable.

The newer applications expose further boundaries. \citet{firoozi2025} implement six guidelines, including option-count limits and ordering options by character count, which differs from SAQUET's numerical or chronological sequencing rule. \citet{arif2024} extend automated review to lecture transcripts with \emph{contextual specificity}, where a question may depend on a diagram visible in the video but absent from the text. Such an item can work within a video yet fail as a standalone question. This demonstrates how rubric applicability, in part, depends on the information available to the learner.

Item format matters equally, such as the case when multiple defensible answers violate a single-best-answer item, but not an intentionally scored multi-select task. Similarly, the use of underscores in code should not automatically trigger a sentence completion flag. Quality assurance should record the item format and intended learner population before applying the rubric, and permit an explicit not applicable or uncertain judgment. Binary outputs without that context can convert legitimate design choices into apparent defects.

\begin{table*}[t]
\centering
\small
\begin{tabularx}{\textwidth}{@{}P{0.205\textwidth}Yc r c P{0.12\textwidth}@{}}
\toprule
\textbf{Criterion} & \textbf{Operational concern} & \textbf{Route} & \textbf{Positive Label ($n$)} & \textbf{F1 range} & \textbf{IRT link} \\
\midrule
Longest option correct & Answer stands out through extra length/detail & T & 59 & .60--1.00 & $b\downarrow,a\downarrow$ \\
Ambiguous information & Wording permits unclear interpretations & N & 40 & .00--.58 & --- \\
Implausible distractors & Foils can be rejected without target knowledge & N & 31 & .00--.86 & --- \\
True/false format & Options reduce to binary truth judgments & T & 27 & .00--1.00 & --- \\
Absolute terms & Unqualified extremes provide a clue & V & 25 & .00--.89 & $a\downarrow$ (Math) \\
Complex/K-type & Responses require combinations of statements & V & 25 & .00--.89 & --- \\
Negative wording & Negation complicates the requested judgment & T & 23 & .00--.91 & $b\uparrow$ \\
Convergence cues & Recombined option components suggest the key & V & 19 & .00--.63 & $b\uparrow$ \\
Lost sequence & Quantities/dates lack an intelligible order & T & 19 & .50--1.00 & $a\uparrow/\downarrow$ \\
Unfocused stem & Stem does not adequately specify the task & V & 16 & .00--.89 & --- \\
None of the above & An exclusion option replaces a substantive answer & T & 13 & .91--1.00 & --- \\
Word repeats & Stem wording selectively points to the key & N & 13 & .53--1.00 & --- \\
Multiple correct answers & A single-best-answer item lacks a unique key & V & 12 & .00--.46 & $b\downarrow,a\downarrow$ \\
Logical cues & Option relations allow unintended deduction & N & 8 & .00--.67 & --- \\
All of the above & A collective option rewards partial recognition & T & 6 & .80--1.00 & $b\downarrow,a\downarrow$ \\
Fill in the blank & Mid-stem gaps interrupt the question & T & 6 & 1.00 & --- \\
Vague terms & Unquantified expressions lack a shared meaning & T & 6 & .00--1.00 & --- \\
Grammatical cues & Only some options grammatically fit the stem & N & 5 & .00--.67 & $b\uparrow$ \\
Gratuitous information & Details add demands unrelated to the task & V & 3 & .00--.50 & --- \\
\bottomrule
\end{tabularx}
\caption{Criterion map anchored in \citet{moore2024}. Routes: T = text/format checks, N = NLP, V = NLP with LLM verification. Positive labels ($n$) count items humans marked as having each flaw, out of 271. F1 ranges span domains (zeros included, uncomputed cells omitted; no pooling). IRT links from \citet{schmucker2025}: $b$ = difficulty, $a$ = discrimination, mixed arrows = domain variation, and --- = no link extracted, not evidence of no association.}
\label{tab:criteria}
\end{table*}

\section{Evidence across the Assessment Workflow}

\subsection{Detection and the accuracy denominator}

SAQUET combines deterministic checks with selective model judgments, while other systems prompt LLMs directly with an item-writing rubric or combine model judgments \citep{moore2023,moore2024,aydin2026,menze2026}. Both approaches are feasible, but comparisons across different datasets cannot establish architectural superiority or isolate benefits of newer models.

SAQUET makes the denominator problem concrete, as its 94.13\% accuracy concerns 5,149 item-criterion decisions, not the fraction of flawed items successfully identified. The published human counts contain 356 positive labels. An always-negative classifier would therefore score $1-356/5149=93.09\%$ accuracy, a baseline calculated here. SAQUET's 38\% exact match, domain micro-F1 of .59--.67, and 75.3\% agreement on a separate acceptability threshold describe different tasks \citep{moore2024}. Table~\ref{tab:criteria} shows why averaging conceals useful distinctions.

Independent medical evidence reinforces the need for more than accuracy. \citet{aydin2026} report strong negative-stem agreement for two models but much weaker alignment with learning objectives. In another approach, two models giving the same answer contrary to the key prompted a review of possible ambiguity. Its specificity was 92.9\%, but sensitivity was only 35.3\% \citep{gerard2026}. Such a signal can focus attention while leaving many defects undetected. A model's failure to solve an item is not itself evidence that the item is defective.

\citet{wu2026} quantify detection within a dental question-generation study. GPT-4 detected 12 of 44 expert-identified flaws across 43 defective items, consisting of 7/7 duplicates, 3/8 flawed distractors, 1/23 off-topic flaws, and 1/6 incorrect or missing answers. This 27.3\% flaw-detection rate is not item-level accuracy, as without defect-free controls, specificity cannot be estimated. Despite requesting revisions, the study reports no separately validated repair-success rate.

Evaluations should report positive class support and per-criterion confusion matrices by domain, with precision, recall, specificity, F1, and uncertainty intervals. Chance corrected agreement is also sensitive to prevalence. Exact match and Hamming loss summarize label completeness and disagreement, but cannot replace criterion-level results. Deliberately flawed examples alone cannot establish specificity on ordinary item banks \citep{sabqat2025}.

In \citet{arif2024}, rules audit 19 criteria, where LLMs check seven IWFs plus contextual specificity on 100 items per model pool. Rules flag about 1.6 flaws per item, LLMs flag 2.18 for GPT-3.5, and 1.86 for GPT-4. These are automated flag counts, not error rates against expert labels. The human rubric also differs, so its 51-65\% flawless item ratings cannot establish relative detector accuracy.

\subsection{Revision and author interaction}

\citet{firoozi2025} combine six rule checks with chained GPT-4o revision and answer-key verification. Their 20-item macroeconomics pilot reports 17 flawless outputs, 18 linguistically acceptable outputs, and 14 preserving the intended content. Two item-writing experts evaluated quality, while a subject expert assessed content alignment. Ratings followed consensus and adjudication of the ordering rule. These are output-quality endpoints, not estimates of detector sensitivity or specificity. An essential point here is that a fluent repair may change what the question tests.

\citet{duesterbeck2026} optimize criterion-specific agent sequences using REINFORCE algorithm and an LLM quality score. Six experts compared two refinements of 30 German biochemistry MCQs, initially generated by Qwen2.5-14B-Instruct to violate quality criteria. Optimized sequences were preferred for stems, but showed no advantage for options. The comparison is with a fixed refinement pipeline, as intentionally flawed inputs limit transfer to ordinary authoring and educator time savings were not measured.

Evidence also varies with the evaluators. In \citet{iembo2026}, ten external evaluators found no significant aggregate difference between original and revised questions, whereas calibrated ratings by four study authors favored the revisions. Calibration and evaluator group changed together, so the difference does not isolate a calibration effect. In \citet{menze2026}, AI-supported authoring was associated with more flags and less interaction. This work suggests fixed task order, self-selection into editing, and imperfect automated labels limit causal interpretation, including attribution to automation bias.

Based on this prior work, we believe studies should preserve original and revised items, distinguish accepted from rejected suggestions, and obtain blinded judgments of whether the intended defect was removed without introducing another. Content correctness, learning-objective alignment, and distractor function need separate checks. Counting generated revisions or repeat tool use does not establish successful remediation.

\subsection{Psychometric and cognitive grounding}

\citet{schmucker2025} connect automated annotations for 7,126 STEM MCQs with IRT parameters estimated from a platform serving 448,000 students. Associations vary by criterion and domain (Table~\ref{tab:criteria}). In the reported regression table, difficulty prediction reaches $R^2=.216$ overall and only .030 in mathematics. A low-difficulty screening example reaches precision .801 with recall .472. These are useful trade-offs for triage, but leave substantial unexplained variation and missed cases.

The association between a flag and difficulty is not the effect of fixing that flag. Subject matter, reading demand, item purpose, and correlated flaws may explain part of the relationship. Moreover, the predictor itself is a fallible automated judgment. New-domain annotation checks are therefore needed before interpreting a regression coefficient as evidence about an underlying flaw. Concept specific IRT calibration also warrants attention when combining parameters from different content areas.

\citet{moore2026cognitive} extends this account by grouping SAQUET's 19 flags into cue-giving, confusion-inducing, and structural-complexity categories. The confusion category comprises ambiguity, unfocused stems, gratuitous information, lost sequence, negative wording, and vague terms. Across three datasets consisting of 653 USMLE, 365 XES3G5M mathematics, and 212 DBE-KT22 database-course items, its presence associates with 14.0\%, 29.4\%, and 18.9\% longer mean response time, respectively, after structural controls. Cue-giving and structural indicators have no significant response-time associations. These are category-level estimates, not validated weights for each constituent criterion.

The difficulty pathway is more conditional. In USMLE, the reported confusion-category indirect estimate through response time is .231, approximately 2.5 times the aggregate estimate of .093. The corresponding category-specific indirect estimates are not significant in XES3G5M and DBE-KT22 uses instructor-assigned ordinal difficulty and a separate attenuation analysis \citep{moore2026cognitive}. Thus, the 2.5-fold result is specific to USMLE. Item-level observational data cannot establish a causal cognitive mechanism, and non-significant interactions do not demonstrate equivalent effects across domains. Machine-translated mathematics text, heterogeneous timing measures, and unvalidated new-domain flags further qualify transfer. The findings motivate targeted review, while controlled comparisons of original and repaired items are still needed to establish measurement benefits.

\subsection{NLP benchmark auditing}

BenchMarker extends educational review to twelve MCQ benchmarks, while separately checking online overlap and answer-option shortcuts \citep{balepur2026}. Its reported tables show how accuracy can conceal weak positive-class performance and how results change between educational and NLP data. For example, Gemini 2.5 Pro obtains writing-flaw F1 of .66 on NLP annotations and .39 on educational annotations. These values concern annotation decisions rather than whole-item correctness.

The benchmark analysis also finds that filtering writing errors changes model rankings beyond random filtering. This makes item construction relevant to the interpretation of NLP evaluations. However, filtered subsets may differ in content and difficulty. Matched-size controls address sampling variation but do not establish a causal effect of correction. Paired repairs, preserved task coverage, and re-evaluation of all models would provide stronger evidence. Online occurrence should likewise be distinguished from demonstrated inclusion in a particular model's training data.

\citet{balepur2026} also evaluated LLM-based repair of TruthfulQA items using detected item-writing flaws as feedback. Across three models, the proportion of items with at least two detected flaws decreased from 97\% before revision to 64–68\% afterward. However, BenchMarker identified newly introduced flaws in 56–68\% of revised items. Because these outcomes were assessed automatically, they do not establish expert-confirmed repair quality or improved measurement. These findings motivate evaluating both the removal of existing defects and the introduction of new ones when assessing automated revision.

\section{Research Priorities}

\paragraph{Make definitions and uncertainty portable.}
Reusable datasets should release exact rubric wording, applicability rules, model versions, and positive and negative examples. Expert annotations should retain disagreements before adjudication and record the intended population and item purpose. Evaluation sets need enough positive cases for each criterion, including semantic defects. Table~\ref{tab:criteria} shows that six criteria have fewer than ten positive cases in the SAQUET reference set, which limits what any F1 value for them can establish. Challenge sets should be reported separately from samples representing deployment prevalence. Shared items across dataset versions should have stable identifiers to prevent leakage and mistaken claims of independent replication.

\paragraph{Prioritize consequences, validate the policy.}
A raw flag count gives a rare grammatical cue and a second correct answer equal weight. Cognitive-category evidence motivates prioritization \citep{moore2026cognitive}, but a review policy must also consider whether a flag is correct, its consequence for the intended assessment, and the cost of checking it. The category results do not validate universal severity weights or imply that cue-giving defects are harmless. Evaluate priorities on held-out domains and independently judged revisions, with student response evidence where appropriate. LLM confidence should not be assumed to be a calibrated probability.

\paragraph{Evaluate the complete human-AI workflow.}
Compare unaided review, flags alone, flags with explanations, and suggested revisions. Measure expert-confirmed defect removal, introduced defects, reviewer time, and escalation. Expert review of unflagged items should estimate missed problems. Subsequent pilots can examine discrimination, intended difficulty, response time, and subgroup performance, connecting authoring efficiency to measurement outcomes.

\paragraph{Design paired item comparisons.}
Paired evaluations should preserve the learning objective while changing the targeted flaw and assign learners to one version only. Shared anchor items can support comparisons on a common scale. Removing a cue may increase difficulty while improving validity, so neither faster responses nor easier items should define success in isolation. Record rejected suggestions and reasons alongside accepted revisions; acceptance alone can reflect misplaced trust.

\section{Limitations}
This review is limited by bounded Crossref retrieval, database coverage, nominated-source selection, and AI-assisted screening without independent replication. The corpus is not exhaustive, and the first author co-authored five of the fourteen included reports, including the toolkit that anchors Table~\ref{tab:criteria}. We did not rerun detectors or reanalyze participant data. Much of the evidence uses small or reused annotation sets, and comparisons span different rubric versions and evaluation units. The cognitive findings are observational and the proposed workflow and consequence-based priorities still require validation.

\section{Conclusion}
AI-enabled IWF review offers interpretable support for screening MCQs, but its evidence is uneven across criteria and stages of use. Surface checks are often tractable, but content sensitive judgments and reliable revision remain harder. The strongest path forward is to evaluate detection, remediation, and assessment outcomes separately and then test how they work together. Rubric flags can organize human attention before deployment, while expert review and response-based validation establish whether the resulting items serve their intended purpose.

% \clearpage
\begingroup
% \hyphenation{assessment}
\bibliography{references}

@article{haladyna2002,
  author = {Haladyna, Thomas M. and Downing, Steven M. and Rodriguez, Michael C.},
  title = {A Review of Multiple-Choice Item-Writing Guidelines for Classroom Assessment},
  journal = {Applied Measurement in Education},
  year = {2002}, volume = {15}, number = {3}, pages = {309--333},
  doi = {10.1207/S15324818AME1503_5}
}

@article{tarrant2006,
  author = {Tarrant, M. and Knierim, A. and Hayes, S. K. and Ware, J.},
  title = {The Frequency of Item Writing Flaws in Multiple-Choice Questions Used in High Stakes Nursing Assessments},
  journal = {Nurse Education Today}, year = {2006}, volume = {26}, number = {8}, pages = {662--671},
  doi = {10.1016/j.nedt.2006.07.006}
}

@article{costello2018,
  author = {Costello, Eamon and Holland, Jane C. and Kirwan, Colette},
  title = {Evaluation of {MCQs} from {MOOCs} for Common Item Writing Flaws},
  journal = {BMC Research Notes}, year = {2018}, volume = {11}, pages = {849},
  doi = {10.1186/s13104-018-3959-4}
}

@inproceedings{moore2023,
  author = {Moore, Steven and Nguyen, Huy A. and Chen, Tianying and Stamper, John},
  title = {Assessing the Quality of Multiple-Choice Questions Using {GPT-4} and Rule-Based Methods},
  booktitle = {Responsive and Sustainable Educational Futures},
  publisher = {Springer}, address = {Cham}, year = {2023}, pages = {229--245},
  url = {https://arxiv.org/abs/2307.08161}
}

@inproceedings{moore2024,
  author = {Moore, Steven and Costello, Eamon and Nguyen, Huy A. and Stamper, John},
  title = {An Automatic Question Usability Evaluation Toolkit},
  booktitle = {Artificial Intelligence in Education},
  publisher = {Springer}, address = {Cham}, year = {2024}, pages = {31--46},
  url = {https://arxiv.org/abs/2405.20529}
}

@inproceedings{firoozi2025,
  author = {Firoozi, Tahereh and Daniels, Lia and Daniels, Vijay and Gierl, Mark},
  title = {An Augmented Intelligence System for Automated Quality Control and Feedback Generation of Multiple Choice Test Items},
  booktitle = {Artificial Intelligence in Education. Posters and Late Breaking Results, Workshops and Tutorials, Industry and Innovation Tracks, Practitioners, Doctoral Consortium, Blue Sky, and WideAIED},
  series = {Communications in Computer and Information Science}, volume = {2592},
  publisher = {Springer}, address = {Cham}, year = {2025}, pages = {86--93},
  doi = {10.1007/978-3-031-99267-4_11}
}

@article{schmucker2025,
  title={The impact of item-writing flaws on difficulty and discrimination in item response theory},
  author={Schmucker, Robin and Moore, Steven},
  journal={Computers and Education: Artificial Intelligence},
  pages={100632},
  year={2026},
  publisher={Elsevier},
  doi = {https://doi.org/10.1016/j.caeai.2026.100632}
}

@inproceedings{balepur2026,
  title={Benchmarker: An education-inspired toolkit for highlighting flaws in multiple-choice benchmarks},
  author={Balepur, Nishant and Rajasekaran, Bhavya and Oh, Hyunjin Jane and Xie, Michael and Desai, Atrey and Gupta, Vipul and Moore, Steven James and Choi, Eunsol and Rudinger, Rachel and Boyd-Graber, Jordan Lee},
  booktitle={Proceedings of the 64th Annual Meeting of the Association for Computational Linguistics (Volume 1: Long Papers)},
  pages={15793--15824},
  year={2026},
  doi = {10.18653/v1/2026.acl-long.719}
}

@article{menze2026,
  author = {Menze, Dennis and Radovi{\'c}, Slavi{\v s}a and Seidel, Niels},
  title = {{AI}-Assisted {MCQ} Creation Increases Item-Writing Flaws through Automation Bias: Evaluation of a Workflow of Multiple {AI} Agents with Teachers},
  journal = {Frontiers in Computer Science}, year = {2026}, volume = {8}, pages = {1831250},
  doi = {10.3389/fcomp.2026.1831250}
}

@article{aydin2026,
  author = {Aydin, Mevl{\"u}t Okan and Co{\c s}kun, Belk{\i}s Nihan and Hamal, {\.{I}}brahim},
  title = {Assessing Multiple-Choice Question Quality in Internal Medicine: A Comparative Analysis of Three Large Language Models against Expert Consensus},
  journal = {Frontiers in Medicine}, year = {2026}, volume = {13}, pages = {1866674},
  doi = {10.3389/fmed.2026.1866674}
}

@article{sabqat2025,
  author = {Sabqat, Mashaal and Khan, Rehan Ahmed and Jawaid, Masood and Sajjad, Madiha},
  title = {Artificial Intelligence Meets Item Analysis ({AI} Meets {IA}): A Study of Chatbot Training and Performance in Detecting and Correcting {MCQ} Flaws},
  journal = {Pakistan Journal of Medical Sciences}, year = {2025}, volume = {41}, number = {3}, pages = {652--656},
  doi = {10.12669/pjms.41.3.11224}
}

@article{iembo2026,
  author = {Iembo, Tatiane and Crist{\'o}v{\~a}o, Helena Landim Gon{\c c}alves and Gon{\c c}alves, Patr{\'i}cia Carla Zanelatto and Montor, Wagner Ricardo and Fucuta, Patr{\'i}cia da Silva and Anbar Neto, Toufic and Andr{\'e}, J{\'u}lio C{\'e}sar and Martins, Milton de Arruda},
  title = {{ChatGPT} as a Tool for Reviewing Multiple-Choice Questions in the Health Sector},
  journal = {Scientific Reports}, year = {2026}, volume = {16}, pages = {21974},
  doi = {10.1038/s41598-026-51988-9}
}

@article{gerard2026,
  author = {G{\'e}rard, Alexandre O. and Merino, Diane and Labriffe, Marc and Rocher, Fanny and Viard, Delphine and Zemori, Laurence and Lavrut, Thibaud and Donker, Erik M. and Pi{\"e}t, Joost D. and Fournier, Jean-Paul and Drici, Milou-Daniel and Destere, Alexandre},
  title = {Evaluating and Leveraging Large Language Models in Clinical Pharmacology and Therapeutics Assessment: From Exam Takers to Exam Shapers},
  journal = {British Journal of Clinical Pharmacology}, year = {2026}, volume = {92}, number = {8}, pages = {2574--2583},
  doi = {10.1002/bcp.70137}
}

@article{duesterbeck2026,
  author = {D{\"u}sterbeck, Lilly Marie and Gr{\"o}ssler, Michael and Credidio, Graziella and Bellmann, Louis and Riemann, Layla Tabea},
  title = {Agent-Based Improvement of Multiple-Choice Question Quality in Medical Education},
  journal = {Studies in Health Technology and Informatics}, year = {2026}, volume = {336}, pages = {2120--2124},
  doi = {10.3233/SHTI260635}
}

@article{wu2026,
  author = {Wu, Jiashun and Ji, Huanzhong and Xie, Yanqing and Kuang, Shijun and Yang, Tao},
  title = {Evaluating the Performance of {GPT-4.0} in Generating and Refining Multiple-Choice Questions for Dental Education},
  journal = {European Journal of Dental Education}, year = {2026},
  doi = {10.1111/eje.70217}
}

@inproceedings{doughty2024,
  author = {Doughty, Jacob and Wan, Zipiao and Bompelli, Anishka and Qayum, Jubahed and Wang, Taozhi and Zhang, Juran and Zheng, Yujia and Doyle, Aidan and Sridhar, Pragnya and Agarwal, Arav and Bogart, Christopher and Keylor, Eric and Kultur, Can and Savelka, Jaromir and Sakr, Majd},
  title = {A Comparative Study of {AI}-Generated ({GPT-4}) and Human-Crafted {MCQs} in Programming Education},
  booktitle = {Proceedings of the 26th Australasian Computing Education Conference},
  year = {2024}, pages = {114--123}, publisher = {Association for Computing Machinery},
  url = {https://arxiv.org/abs/2312.03173}
}

@article{vanderlee2021,
  author = {van der Lee, Chris and Gatt, Albert and van Miltenburg, Emiel and Krahmer, Emiel},
  title = {Human Evaluation of Automatically Generated Text: Current Trends and Best Practice Guidelines},
  journal = {Computer Speech \& Language}, year = {2021}, volume = {67}, pages = {101151},
  doi = {10.1016/j.csl.2020.101151}
}

@inproceedings{arif2024,
  author = {Arif, Taimoor and Asthana, Sumit and Collins-Thompson, Kevyn},
  title = {Generation and Assessment of Multiple-Choice Questions from Video Transcripts using Large Language Models},
  booktitle = {Proceedings of the Eleventh ACM Conference on Learning @ Scale},
  year = {2024}, pages = {530--534},
  publisher = {Association for Computing Machinery},
  doi = {10.1145/3657604.3664714}
}

@inproceedings{moore2026cognitive,
  author = {Moore, Steven},
  title = {The Cognitive Cost of Poor Item Design: How Item-Writing Flaws Shape Response Time and Difficulty Across Assessment Domains},
  booktitle = {Proceedings of the 19th International Conference on Educational Data Mining},
  publisher = {International Educational Data Mining Society},
  address = {Seoul, Republic of Korea}, year = {2026}, pages = {293--305},
  doi = {10.5281/zenodo.21040093}
}

@article{gusenbauer2024beyond,
  title={Beyond Google Scholar, Scopus, and Web of Science: An evaluation of the backward and forward citation coverage of 59 databases' citation indices},
  author={Gusenbauer, Michael},
  journal={Research synthesis methods},
  volume={15},
  number={5},
  pages={802--817},
  year={2024},
  publisher={Wiley Online Library},
  doi = {10.1002/jrsm.1729}
}

@article{gottlieb2023educator,
  title={Educator's blueprint: A how-to guide for developing high-quality multiple-choice questions},
  author={Gottlieb, Michael and Bailitz, John and Fix, Megan and Shappell, Eric and Wagner, Mary Jo},
  journal={AEM Education and Training},
  volume={7},
  number={1},
  pages={e10836},
  year={2023},
  publisher={Wiley Online Library},
  doi = {10.1002/aet2.10836}
}

@inproceedings{balepur2025these,
  title={Which of these best describes multiple choice evaluation with llms? a) forced b) flawed c) fixable d) all of the above},
  author={Balepur, Nishant and Rudinger, Rachel and Boyd-Graber, Jordan Lee},
  booktitle={Proceedings of the 63rd Annual Meeting of the Association for Computational Linguistics (Volume 1: Long Papers)},
  pages={3394--3418},
  year={2025},
  doi={10.18653/v1/2025.acl-long.169}
}
\endgroup
\end{document}